\documentclass[10pt, a4paper]{article}

\usepackage[final]{lrec2026} 

\usepackage{algorithm}
\usepackage{algorithmic}
\usepackage{booktabs} 
\usepackage{amsmath}
\usepackage{bookmark}
\usepackage{hyperref}
 
\title{CLASE: A Hybrid Method for Chinese Legalese Stylistic Evaluation}

\name{Yiran Rex Ma$^\dagger$\thanks{$^\dagger$Equal contribution.}\textsuperscript{\rm 1}, Yuxiao Ye$^\dagger$\textsuperscript{\rm 2}, Huiyuan Xie\textsuperscript{\rm 2}} 

\address{\textsuperscript{\rm 1}School of Foreign Languages, Peking University\\
\textsuperscript{\rm 2}Department of Computer Science and Technology, Tsinghua University\\
         Beijing, China \\
         \texttt{yiranrexma@outlook.com}\\}

\abstract{
  Legal text generated by large language models (LLMs) can usually achieve reasonable factual accuracy, but it frequently fails to adhere to the specialised stylistic norms and linguistic conventions of legal writing. In order to improve stylistic quality, a crucial first step is to establish a reliable evaluation method. However, having legal experts manually develop such a metric is impractical, as the implicit stylistic requirements in legal writing practice are difficult to formalise into explicit rubrics. Meanwhile, existing automatic evaluation methods also fall short: reference-based metrics conflate semantic accuracy with stylistic fidelity, and LLM-as-a-judge evaluations suffer from opacity and inconsistency. To address these challenges, we introduce \textit{\textbf{CLASE}} (\textbf{C}hinese \textbf{L}eg\textbf{A}lese \textbf{S}tylistic \textbf{E}valuation), a hybrid evaluation method that focuses on the stylistic performance of legal text. The method incorporates a hybrid scoring mechanism that combines 1) linguistic feature-based scores and 2) experience-guided LLM-as-a-judge scores. Both the feature coefficients and the LLM scoring experiences are learned from contrastive pairs of authentic legal documents and their LLM-restored counterparts. This hybrid design captures both surface-level features and implicit stylistic norms in a transparent, reference-free manner. Experiments on 200 Chinese legal documents show that CLASE achieves substantially higher alignment with human judgments than traditional metrics and pure LLM-as-a-judge methods. Beyond improved alignment, CLASE provides interpretable score breakdowns and suggestions for improvements, offering a scalable and practical solution for professional stylistic evaluation in legal text generation$^\star$\thanks{$^\star$Code and data for CLASE is available at: \url{https://github.com/rexera/CLASE}.}. 
 \\ \newline  }

\begin{document}

\maketitleabstract

\section{Introduction}

Large Language Models (LLMs) have significantly advanced legal content generation, with advances on legal reasoning tasks and bar examinations \cite{bommarito2022gpt}. These developments have generated considerable interest in automating legal document production, ranging from contract drafting to judicial decision writing. However, while LLMs excel at generating factually accurate and logically coherent legal content, they consistently struggle with the specialized stylistic conventions (``legalese'') that characterize professional legal discourse \cite{tiersma1999legal,courtWritingGuide2010}.

\begin{figure}
    \centering
    \includegraphics[width=1\linewidth]{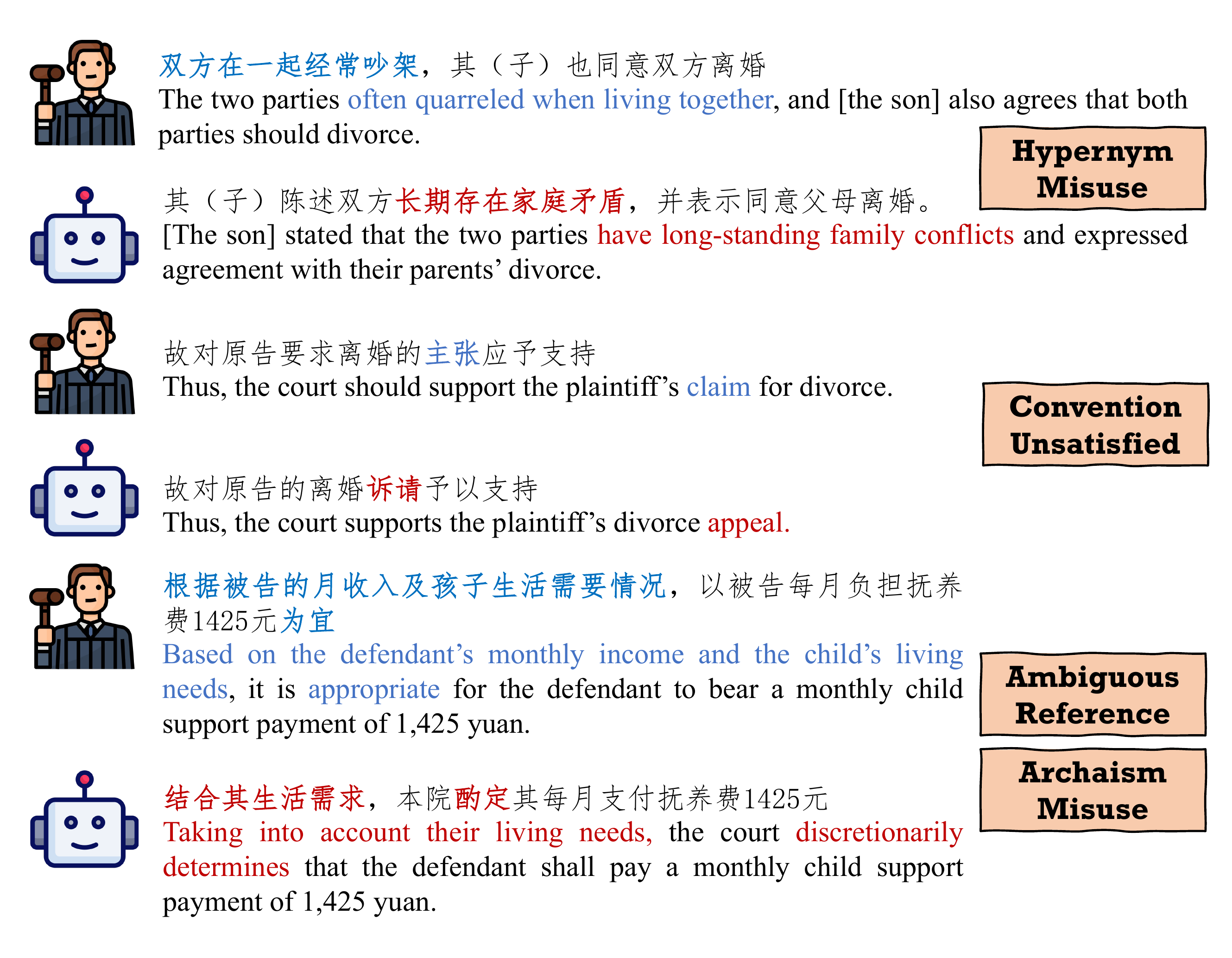}
    \caption{Comparison of Authentic Legal Writing and LLM-Generated Counterpart.}
    \label{fig:intro}
\end{figure}

Legal writing demands adherence to established stylistic norms that extend far beyond semantic correctness. Professional legal documents must conform to specific collocation patterns, formal register requirements, and domain-specific linguistic conventions that signal authority and credibility within the legal community and facing the general public \cite{foley2002language,li2020cohesion,lu2021legal,sun2017linguistic,li2022lexical}. 

LLM-generated legal documents exhibit two primary categories of stylistic deficiencies: \textit{insufficient sophistication}, failing to meet the expectations through inappropriate colloquialisms or non-standard term choices; and \textit{excessive stylistic elaboration}, where models artificially create a formal impression through verbose, unnecessarily complex, or archaic constructions. This overcompensation often results in hallucinated legal concepts. These dual deficiencies risk undermining professional acceptability (see Figure \ref{fig:intro}).

Current evaluation for legal text generation focuses on \textbf{reasoning capabilities} while neglecting \textbf{stylistic quality assessment} \cite{chalkidis2022lexglue,zhong2018cail2018,fu2022jec}. This assumes that semantic accuracy, represented by factual correctness, logical argumentation accuracy, and legal knowledge demonstration, constitutes the primary criterion for legal text quality, whereas stylistic appropriateness is just as important in legal competence.

\textbf{Stylistic quality} in legal writing encompasses multiple intricate dimensions that traditional evaluation approaches fail to capture. These include: 1) precise diction and lexical choice; 2) adherence to linguistic/legal conventions; 3) appropriate lexical collocation patterns; and 4) domain-specific common sense and world knowledge \cite{foley2002language}. These elements collectively contribute to what legal practitioners recognize as ``professional'' or ``sophisticated'' writing style, yet they remain largely underexplored in current automated evaluation systems. 

Traditional natural language generation (NLG) metrics could be inadequate for capturing stylistic nuances in legal text. Both branches of n-gram and embedding-based methods conflate semantic similarity with stylistic appropriateness \cite{mellish1998evaluation,papineni2002bleu,zhang2019bertscore}. These metrics may award high scores to texts that preserve semantic content while exhibiting stylistic violations that would be immediately apparent to legal professionals. Recent advances in LLM-as-a-Judge evaluation \cite{chan2023chateval} offer intuitive solutions through instruction understanding, but 1) suffer from limited interpretability \cite{wang2023survey}; 2) exhibit biases and consistency issues that undermine their reliability.

The challenge is compounded by the implicit nature of legal stylistic expertise. Writing appropriateness relies on tacit professional knowledge that resists explicit formalization. Legal professionals develop stylistic intuition through years of practice and exposure to exemplary texts, making it difficult to translate this expertise into comprehensive evaluation rubrics. Manual annotation by legal experts is prohibitively time-consuming for large-scale evaluation needs.

To address these gaps, we introduce CLASE\footnote{With inspiration from Spanish ``\textit{con clase}'' (having class).} (\textbf{C}hinese \textbf{L}eg\textbf{A}lese \textbf{S}tylistic \textbf{E}valuation), a hybrid evaluation framework designed specifically for assessing stylistic fidelity in legal text generation. Our approach combines objective linguistic feature analysis with experience-guided LLM evaluation, addressing the limitations of existing methods while maintaining interpretability and reference-free operation. The framework employs contrastive learning using authentic legal documents and their stylistically-restored counterparts, enabling automatic acquisition of evaluation criteria that reflect actual professional expectations without requiring manual annotation. CLASE operates in three phases: 1) automated synthesization of contrastive training pairs; 2) training-free contrastive learning that builds experience pools; and 3) hybrid scoring combining linguistically-grounded objective measures with subjectively-informed LLM assessment. 

Our primary contributions include:

\begin{itemize}
\item \textbf{A novel hybrid evaluation architecture} that  addresses the neglected dimension of stylistic quality in legal text generation, combining objective linguistic feature analysis with experience-guided LLM evaluation to provide comprehensive stylistic assessment while maintaining interpretability and reference-free operation.

\item \textbf{A contrastive learning framework} that eliminates expensive manual annotation requirements while ensuring alignment with professional standards, employing authentic legal documents and their deliberately stylistically-degraded counterparts to automatically acquire evaluation criteria that capture actual model defects and reflect actual professional expectations.

\item \textbf{Empirical validation} demonstrating  improved correlation with human expert judgments compared to traditional metrics and pure LLM-based approaches, with interpretable, actionable natural language feedback with improvement strategies.
\end{itemize}

CLASE addresses a gap in domain-specific, stylistics-oriented evaluation, providing a scalable and transparent solution for professional stylistic assessment. The core principles can extend to other domains requiring specialized stylistic conformity, offering a general approach to stylistic evaluation in professional text generation.

\section{Related Work}

\subsection{NLG Evaluation}

NLG evaluation operates across three primary paradigms: \textit{n-gram based lexical metrics}, \textit{neural embedding approaches}, and \textit{LLM-as-a-Judge systems}. 1) Reference-based, n-gram lexical metrics such as BLEU \cite{papineni2002bleu}, ROUGE \cite{lin2004rouge}, and METEOR \cite{banerjee2005meteor} measure lexical similarity between generated text and human-written references, but these approaches face challenges in capturing semantic adequacy and stylistic appropriateness \cite{reiter2009validity,novikova2017need}. 2) Embedding-based neural metrics like BERTScore \cite{zhang2019bertscore} and MoverScore \cite{zhao2019moverscore} leverage contextual representations to improve correlation, though they measure semantic distance outside of formal dimensions \cite{schmidt2021large}. 3) LLM-as-a-Judge paradigm \cite{liu2023geval,chan2023chateval} offers reference-free assessment capabilities but introduces challenges including self-enhancement bias, positional bias, and interpretability concerns \cite{zheng_judging_2023,li_leveraging_2024}. Other reference-free evaluation methods have emerged to address reference scarcity \cite{ito_reference-free_2025}, employing techniques such as learning from human judgments through regression models \cite{rei2021references}, similarity-based approaches, and pseudo-rating methods. However, they often require substantial training and face challenges in ensuring consistent evaluation criteria.

\subsection{Legal Text Generation and Evaluation}

Early research focused on adapting general-purpose models through fine-tuning on legal corpora \cite{chalkidis2020legal}, leading to dedicated legal LLMs including LawGPT \cite{zhou2024lawgpt}, ChatLaw \cite{cui2023chatlaw}, and Lawyer-LLaMA \cite{huang2023lawyer}. These models demonstrate improvements on legal tasks compared to general models, with applications spanning document summarization \cite{deroy2023ready}, case analysis \cite{savelka2023explaining}, and legal question answering \cite{fu2022jec}. As for benchmarking, LawBench \cite{fei2023lawbench} provides multi-dimensional assessment across memorization, understanding, and application levels. LegalEval-Q \cite{li2025legaleval} focuses on clarity, coherence, and terminology. Chinese legal benchmarks including LAiW \cite{dai2023laiw}, UCL-Bench \cite{gan2025ucl}, JuDGE \cite{su2025judge}, and CaseGen \cite{li2025casegen} offer evaluation covering various capability levels and practical applications. Gap analysis research \cite{hou2024gaps,ma-2025-androids} has identified issues in LLM-generated reasoning/analysis, highlighting the need for fine-grained evaluation, such as hybrid approaches combining automated metrics with expert human judgment \cite{guha2023legalbench,shao2025legal}.

\subsection{Style and Stylistic Evaluation}

Prevalent perspective in computational linguistics outlines ``style'' broadly as extents of formality, politeness, simplicity, personality, emotion, etc. \cite{jin_deep_2022}, focusing on style transfer (periods, genre, authors...), authorship attribution/stylistic fingerprints of LLMs \cite{bitton2025detectingstylisticfingerprintslarge}, and stylometric analysis \cite{juola2006authorship,argamon2003stylistic}. Recent advances in content-independent style embeddings address content leakage challenges. StyleDistance \cite{styledistance2024} employs LLM-synthetic contrastive parallel text to create controlled stylistic variations to train separate embeddings, which gives us crucial insights into addressing stylistic quality assessment. Our work focuses on stylistic \textit{quality} assessment, which differs from classical style transfer/analysis. We assess adherence to domain-specific conventions and professional standards in Chinese legal contexts.

\section{Method}

\begin{figure}[ht]
  \centering
  \includegraphics[width=1\linewidth]{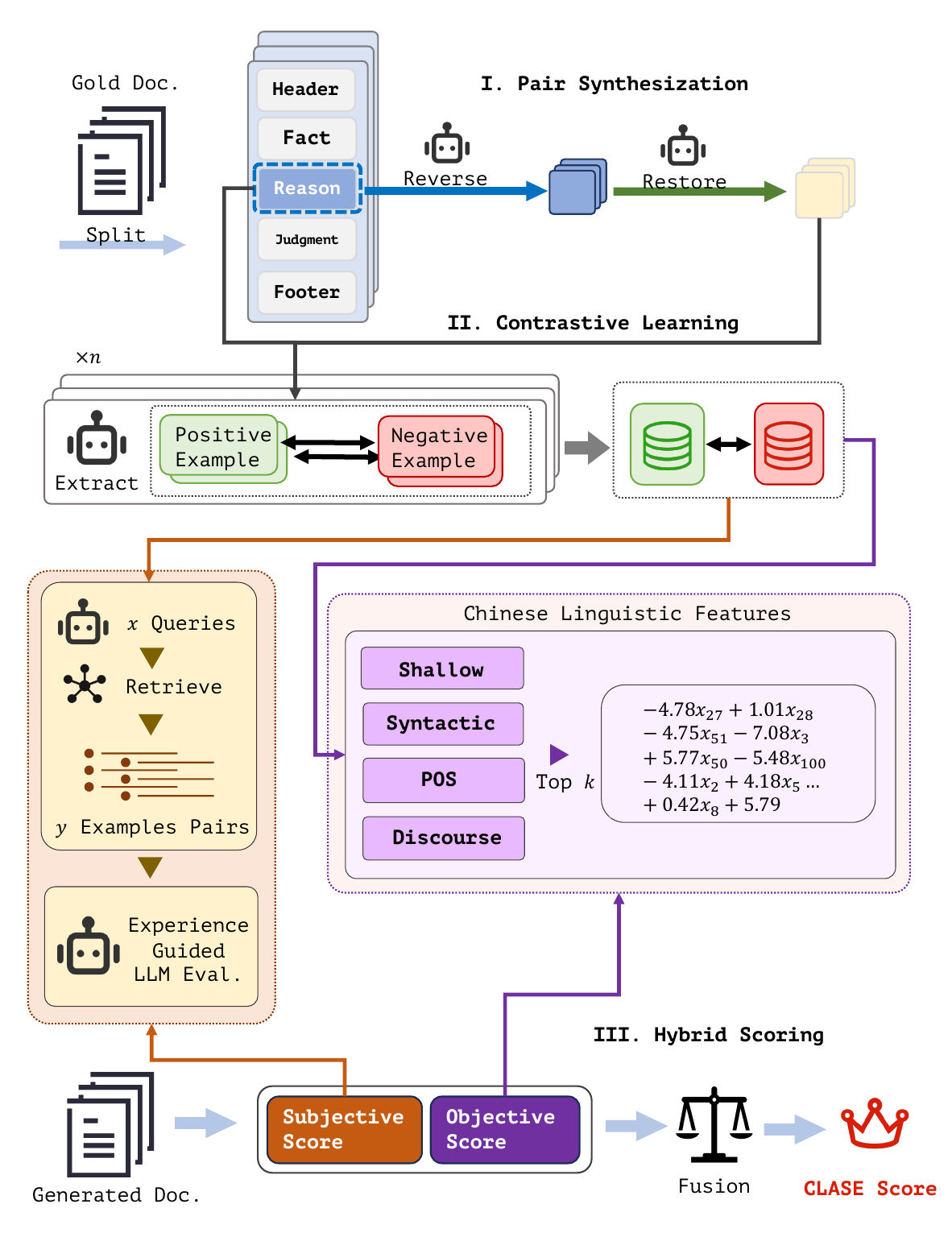}
  \caption{CLASE Overview: (I) contrastive pair synthesization from authentic legal documents; (II) training-free contrastive learning to build positive/negative example pools; (III) hybrid scoring combining objective linguistic features with experience-guided LLM evaluation.}
  \label{fig:clase_overview}
\end{figure}

CLASE adopts a three-stage approach (see Figure \ref{fig:clase_overview}): \textit{contrastive pair synthesization}, \textit{training-free contrastive learning}, and \textit{hybrid scoring}. The framework requires no manual annotations while capturing both surface-level linguistic patterns and implicit stylistic norms.

\subsection{Contrastive Pair Synthesization}

We construct training exemplars from authentic Chinese judgment documents\footnote{Following professional practice, we list five sections in first-instance civil judgments: 1) header (case and party information), 2) facts (claims and findings), 3) reasoning (dispute analysis and rationale), 4) judgment (legal basis and outcome), and 5) footer (appeal information and signatures). This constitutes ``split'' in Figure \ref{fig:clase_overview}.}, \textbf{focusing on the reasoning sections} where stylistic quality most critically impacts professional acceptability. For each original text segment $t_{\text{gold}}$, we generate contrastive learning pairs through a two-stage, prompt-guided transformation: $t_{\text{reverse}} = \pi_1(t_{\text{gold}})$ and $t_{\text{restored}} = \pi_2(t_{\text{reverse}})$, where $\pi_1$ performs stylistic degradation by converting legalese to colloquial expression while preserving semantics, named entities, and topic chains, and $\pi_2$ attempts restoration from the degraded text back to legalese. 

\subsection{Training-Free Contrastive Learning}

This stage operates through structured steps to accumulate stylistic knowledge without manual annotations. Each learning step $\tau^{(i)}$ processes a contrastive pair $(t_{\text{gold}}^{(i)}, t_{\text{restored}}^{(i)})$ to extract labeled exemplars for regression.

For the $i$-th learning step $\tau^{(i)}$, we define the input as a single contrastive pair $(t_{\text{gold}}^{(i)}, t_{\text{restored}}^{(i)})$. The goal of each learning step is to identify stylistic issues and extract corresponding positive-negative exemplar pairs. Within each step, we perform guided comparison to identify a set of stylistic problems $\mathcal{I}^{(i)} = \{I_1^{(i)}, I_2^{(i)}, ..., I_{m_i}^{(i)}\}$, where each $I_j^{(i)}$ represents a specific stylistic issue discovered by comparing the gold and restored texts:

$$\mathcal{I}^{(i)} = \pi_{\text{identify}}(t_{\text{gold}}^{(i)}, t_{\text{restored}}^{(i)})$$

where $\pi_{\text{identify}}$ outputs structured issue descriptions. For each identified problem $I_j^{(i)} \in \mathcal{I}^{(i)}$, we extract a positive-negative exemplar pair $(e_{\text{pos}}^{(i,j)}, e_{\text{neg}}^{(i,j)})$ through a prompt-constrained \textbf{one-to-one correspondence} $\Theta: e_{\text{pos}}^{(i,j)} \leftrightarrow e_{\text{neg}}^{(i,j)}$, ensuring paired positions in both pools address the same stylistic aspect. The extracted exemplars are accumulated into two separate pools: positive experience pool $\mathcal{P}_{\text{pos}}$ and negative experience pool $\mathcal{P}_{\text{neg}}$. 

After completing $N$ learning steps, we perform logistic regression on the accumulated experience pools. Inspired by the method in \citet{huang_exploring_2018}'s Chinese readability assessment, we extract linguistic features $F(e) = \{f_1, f_2, ..., f_{100}\}$ from each exemplar $e$ in both pools, encompassing surface-level characteristics including character complexity, part-of-speech distributions, syntactic patterns, and discourse markers, among others\footnote{Refer to \citet{huang_exploring_2018} or CLASE repository (\url{https://github.com/rexera/CLASE}) for a comprehensive list of features.}. Features are z-score normalized to ensure commensurate scales before coefficient comparison. The regression model learns to distinguish positive from negative exemplars:

$$P(\text{positive}|F(e)) = \sigma(\mathbf{w}^T F(e) + b)$$

where $\mathbf{w}$ represents learned feature weights and $\sigma$ is the sigmoid function. This provides a feature-based scoring mechanism that generalizes from the accumulated exemplars to evaluate new texts.

\begin{algorithm}
\caption{Training-Free Contrastive Learning}
\begin{algorithmic}
\STATE \textbf{Input:} Document pairs $\{(t_{\text{gold}}^{(i)}, t_{\text{restored}}^{(i)})\}_{i=1}^{N}$
\STATE \textbf{Output:} Experience pools $\mathcal{P}_{\text{pos}}$, $\mathcal{P}_{\text{neg}}$, coefficients $\mathbf{w}$
\STATE 
\STATE $\mathcal{P}_{\text{pos}} \leftarrow \emptyset$, $\mathcal{P}_{\text{neg}} \leftarrow \emptyset$
\FOR{$i = 1$ to $N$} 
    \STATE $\mathcal{I}^{(i)} \leftarrow \pi_{\text{identify}}(t_{\text{gold}}^{(i)}, t_{\text{restored}}^{(i)})$ 
    \FOR{$j = 1$ to $|\mathcal{I}^{(i)}|$}
        \STATE $(e_{\text{pos}}^{(i,j)}, e_{\text{neg}}^{(i,j)}) \leftarrow \Phi(I_j^{(i)})$ 
        \STATE $\mathcal{P}_{\text{pos}} \leftarrow \mathcal{P}_{\text{pos}} \cup \{e_{\text{pos}}^{(i,j)}\}$, $\mathcal{P}_{\text{neg}} \leftarrow \mathcal{P}_{\text{neg}} \cup \{e_{\text{neg}}^{(i,j)}\}$ \COMMENT{$\Theta$ correspondence}
    \ENDFOR
\ENDFOR
\STATE $X \leftarrow$ [FeatureExtract($e$) for $e \in \mathcal{P}_{\text{pos}} \cup \mathcal{P}_{\text{neg}}$]
\STATE $y \leftarrow$ [1 for $e \in \mathcal{P}_{\text{pos}}$] + [0 for $e \in \mathcal{P}_{\text{neg}}$]
\STATE $\mathbf{w} \leftarrow$ LogisticRegression($X$, $y$)
\STATE \textbf{return} $\mathcal{P}_{\text{pos}}$, $\mathcal{P}_{\text{neg}}$, $\mathbf{w}$
\end{algorithmic}
\end{algorithm}

\subsection{Hybrid Scoring}

CLASE produces a final score $\Psi(t)$ through hybrid combination of objective (\textit{CLASE Obj}) and subjective (\textit{CLASE Subj}) assessments. Without reference, for each generated text $t$, it eventually offers a sigmoid-fused, $[0,1]$ score from both linguistic features and experience-guided LLM judging.

Given input text $t$, \textit{CLASE Obj} extracts linguistic features $F(t) = \{f_1, f_2, ..., f_{100}\}$ and select the top-$k$ features based on logistic regression coefficient magnitudes. The objective score is computed as:

$$\Psi_{obj}(t) = 10 \times \sigma(\mathbf{w}^T F_k(t))$$

where $F_k(t)$ represents the selected $k$ features, $\mathbf{w}$ contains learned regression weights, and the output is normalized to $[0,10]$.

\textit{CLASE Subj} evaluates seven dimensions based on legal writing practice through retrieval-augmented, experience-guided assessment. 

We define the dimension set $\mathcal{D} = \{d_1, d_2, ..., d_7\}$ with respective weights: noun usage (30\%), verb usage (30\%), adjective usage (20\%), function words (5\%), sentence coherence (5\%), sentence structure (5\%), and collocations (5\%). 

For each dimension $d \in \mathcal{D}$, 1) LLM judge $\pi$ analyzes the input text $t$ and generates $x$ queries focusing on potential stylistic issues within dimension $d$. 2) for each query, we retrieve top-$y$ negative exemplars from $\mathcal{P}_{\text{neg}}$ and obtain their corresponding positive exemplars through $\Theta$. 3) $\pi$ score the text in $[0,10]$ using these contrastive exemplar pairs as contextual guidance.

The final CLASE score combines both components through empirically-calibrated hybrid fusion. We use equal weighting (0.5 each) based on pilot studies showing optimal performance at this balance. The sigmoid transformation ensures output normalization while preserving relative rankings:

$$\Psi(t)' = 0.5 \times \Psi_{obj}(t) + 0.5 \times \Psi_{subj}(t)$$
$$\Psi(t) = \frac{1}{1 + \exp(-10 \times (\Psi(t)'/10 - 0.5))}$$

\begin{algorithm}
\caption{Hybrid Scoring}
\label{alg}
\begin{algorithmic}
\STATE \textbf{Input:} Text $t$, experience pools $\mathcal{P}_{\text{pos}}, \mathcal{P}_{\text{neg}}$, weights $\mathbf{w}$, dimension set $\mathcal{D}$, $k$, query count $x$, retrieval count $y$
\STATE \textbf{Output:} Final CLASE score $\Psi(t)$
\STATE // Objective Scoring
\STATE $F_k(t) \leftarrow$ ExtractTopKFeatures($t$, $k$)
\STATE $\Psi_{obj}(t) \leftarrow 10 \times \sigma(\mathbf{w}^T F_k(t))$
\STATE // Subjective Scoring
\STATE $\Psi_{subj}(t) \leftarrow 0$
\FOR{$d \in \mathcal{D}$}
    \STATE $\mathcal{A}_d \leftarrow \pi$.AnalyzeText($t$, $d$) \COMMENT{$\pi$ identifies potential issues in dimension $d$}
    \STATE $Q \leftarrow \pi$.GenerateQueries($\mathcal{A}_d$, $x$) 
    \STATE $\mathcal{E} \leftarrow \emptyset$
    \FOR{$q_i \in Q$}
        \STATE $N_{q_i} \leftarrow$ TopKSimilar($q_i$, $\mathcal{P}_{\text{neg}}$, $y$) \COMMENT{Retrieve $y$ similar negatives}
        \STATE $P_{q_i} \leftarrow$ GetCorresponding($N_{q_i}$, $\mathcal{P}_{\text{pos}}$) \COMMENT{Get paired positives via $\Theta$}
        \STATE $\mathcal{E} \leftarrow \mathcal{E} \cup \{(P_{q_i}, N_{q_i})\}$
    \ENDFOR
    \STATE $\Psi_{subj}(t, d) \leftarrow \pi$.Evaluate($t$, $\mathcal{E}$, $d$)
    \STATE $\Psi_{subj}(t) \leftarrow \Psi_{subj}(t) + \beta_d \cdot \Psi_{subj}(t, d)$
\ENDFOR
\STATE $\Psi(t) \leftarrow$ SigmoidFusion($\Psi_{obj}(t)$, $\Psi_{subj}(t)$)
\STATE \textbf{return} $\Psi(t)$
\end{algorithmic}
\end{algorithm}

\section{Experiments}

\subsection{Experimental Setup}

We build, train, and test CLASE on Qwen-2.5 model family \cite{qwen25_2024}. \textbf{At learning time}, we conduct experiments using 4000 Chinese civil judgment documents (learning step $N=4000$)\footnote{https://wenshu.court.gov.cn}. For the contrastive pair synthesization phase, we employ the 7B model for stylistic degradation ($\pi_1$), 32B for restoration ($\pi_2$), and 72B for experience pool construction. 

\textbf{At test time}, we sample 200 additional documents from the same data source, outside the training scope. We follow the same pipeline to generate colloquially reverse versions, then employ GPT-4o \cite{openai2024gpt4o} to create restored versions with controlled variations to simulate different legalese proficiency levels and varied emphasis on legal writing requirements: efficiency, thoroughness, structure, formality, educational\footnote{Legal writing serves as not only an instrument for the rule of law, judicial practices, and law enforcement, but also a vital medium for shaping public legal awareness. Detailed implementation is in \href{https://github.com/rexera/CLASE}{our repository}.}. 

The top-$k$ feature selection uses absolute coefficient values from L2-regularized logistic regression. Text segmentation follows jieba tokenization with character-level boundary detection for span alignment. For the subjective scoring component, we use the 72B model with a naive Chain-of-Thought (CoT) prompting \cite{wei_chain--thought_2023} as the judge model. For retrievals, we use embeddings of the 7B model with cosine similarity as the distance metric. In the main experiments, we configure the retrieval parameters as query count $x=10$ and retrieval count $y=10$, with the objective component using top $k=25$  features. All experiments are conducted on 8 NVIDIA A800-SXM4-80GB GPUs with vLLM \cite{kwon2023efficient} for model inference.

\subsection{Evaluation}

We recruit two legal domain experts to conduct evaluation for 200 restored documents across the seven aforementioned stylistic dimensions and respective weights. Each expert independently assigns scores from 0-10 for each dimension comparing gold and restored documents. Inter-annotator agreement achieves Krippendorff's alpha of 0.72 \cite{krippendorff2011computing}, indicating reliability. We evaluate system performance using Pearson correlation coefficient $r$ \cite{pearson1895note}, Spearman rank correlation $\rho$ \cite{spearman1904proof}, and Kendall's $\tau$ \cite{kendall1938new}.

\subsection{Baselines}

1) Traditional reference-based metrics include standard n-gram methods (character-level F1, BLEU, ROUGE, METEOR) and embedding-based semantic similarity (BERTScore). The F1 baseline uses character-level exact matching with precision-recall harmonic mean. 2) LLM-as-a-Judge methods employ GPT-4o-mini and Qwen-2.5-72B in both reference-based (``-ref'') and reference-free configurations, evaluating across the same seven stylistic dimensions with 0-10 scoring. 3) CLASE variants include subjective-only (CLASE-Subj), objective-only (CLASE-Obj), and hybrid fusion (CLASE-Mix).

\section{Results and Analysis}

\begin{table}[htbp]
\centering
\caption{Main Correlation Results}
\label{tab:main_weighted}
\begin{tabular}{llll}
\toprule
Method &  $r$ &  $\rho$ &  $\tau$ \\
\midrule
ROUGE & 0.4250 & 0.4807 & 0.3355 \\
BLEU & 0.4160 & 0.3383 & 0.2306 \\
F1 & 0.3907 & 0.3214 & 0.2213 \\
BERTScore & 0.3145 & 0.2010 & 0.1378 \\
METEOR & 0.3586 & 0.2949 & 0.2020 \\
GPT-4o-mini-ref & 0.3206 & 0.3061 & 0.2083 \\
Qwen2.5-72B-ref & 0.3049 & 0.2482 & 0.1808 \\
GPT-4o-mini & 0.0689 & 0.0747 & 0.0528 \\
Qwen2.5-72B & 0.2416 & 0.2336 & 0.1855 \\
CLASE-Subj & 0.3165 & 0.3063 & 0.2148 \\
CLASE-Obj & 0.7923 & 0.7568 & 0.5692 \\
\textbf{CLASE-Mix} & \textbf{0.8271} & \textbf{0.8109} & \textbf{0.6180} \\
\bottomrule
\end{tabular}
\end{table}

\begin{table}[htbp]
    \centering
    \caption{Variance and Dispersion Analysis}
    \label{tab:variance_analysis}
    \begin{tabular}{lrrr}
    \toprule
    Method & Std & Variance & CV \\
    \midrule
    \textbf{CLASE-Mix} & \textbf{2.021} & \textbf{4.086} & \textbf{0.365} \\
    Human & 1.722 & 2.965 & 0.383 \\
    Qwen2.5-72B-ref & 0.639 & 0.408 & 0.085 \\
    GPT-4o-mini-ref & 0.574 & 0.330 & 0.082 \\
    CLASE-Subj & 0.473 & 0.223 & 0.079 \\
    GPT-4o-mini & 0.473 & 0.223 & 0.062 \\
    ROUGE & 0.363 & 0.132 & 0.669 \\
    Qwen2.5-72B & 0.176 & 0.031 & 0.022 \\
    CLASE-Obj & 0.118 & 0.014 & 2.695 \\
    METEOR & 0.103 & 0.011 & 0.231 \\
    BLEU & 0.088 & 0.008 & 0.385 \\
    F1 & 0.072 & 0.005 & 0.120 \\
    BERTScore & 0.035 & 0.001 & 0.043 \\
    \bottomrule
    \end{tabular}
    \end{table}

\begin{figure}
    \centering
    \includegraphics[width=1\linewidth]{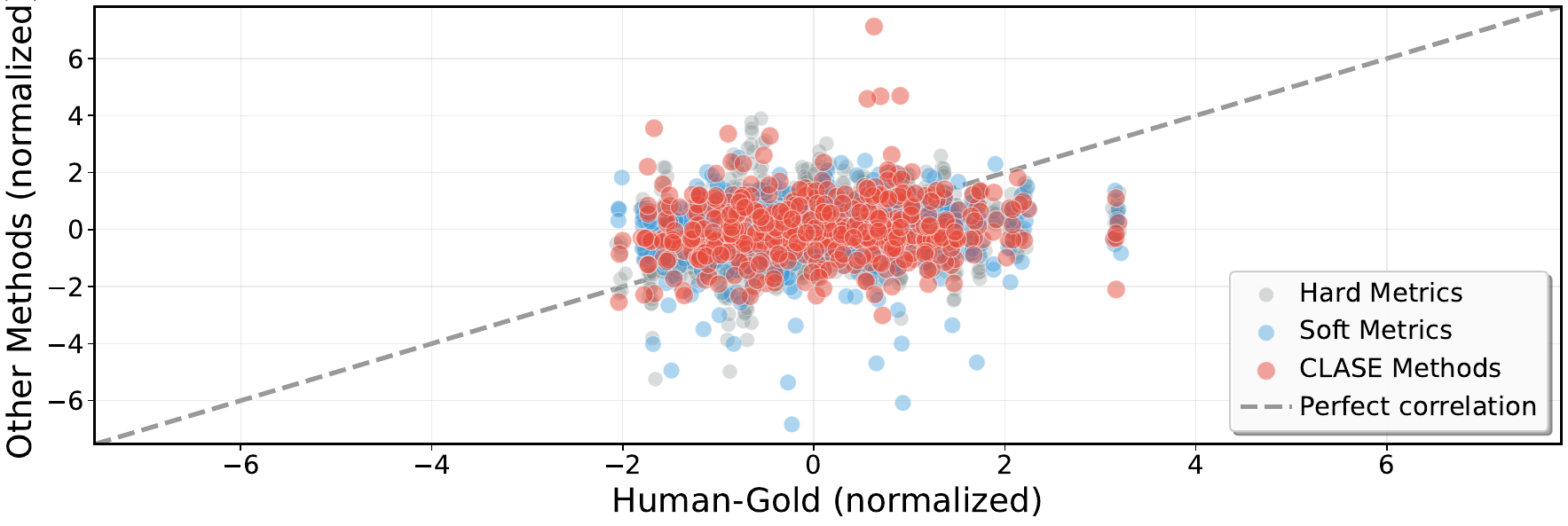}
    \caption{Correlation analysis between evaluation methods and human judgments. Points closer to the diagonal line indicate better alignment with human evaluation.}
    \label{fig:variance_analysis}
\end{figure}

\subsection{Main Results}

\paragraph{Correlation} As shown in Table \ref{tab:main_weighted}, 1) hard metrics fail to capture stylistic variation. These metrics focus on surface-level lexical matching or semantic similarity, which is inadequate for capturing stylistic features or differentiating documents based on stylistic variation alone. 2) LLM-as-a-judge methods (soft metrics) have inherent limitations in providing fine-grained numerical scores for style. It is better suited for generating qualitative, natural language feedback and could benefit from the grounding provided by objective measures. 3) Notably, CLASE-Obj achieves strong correlation. This suggests that the contrastive learning approach successfully identifies useful linguistic features that correlate with professional legalese. When fused with CLASE-Subj, it gains incremental improvement for introducing flexible nature of LLMs. 

Figure \ref{fig:variance_analysis} visualizes this alignment through normalized scores, where the x-axis represents standardized human judgments (z-scores) and the y-axis shows standardized scores from each evaluation method. The diagonal line $y=x$ indicates perfect correlation. CLASE effectively cluster closer to the diagonal line compared to hard and soft metrics, indicating superior capability in capturing human-perceived stylistic quality. 

\paragraph{Dispersion} Table \ref{tab:variance_analysis} reveals critical evaluation characteristics through the coefficient of variation (CV), which normalizes variability by mean scores to enable fair comparison. Traditional metrics exhibit extremely low CV values, whereas LLM-as-a-Judge methods show relatively higher values yet still more ``conservative'' than human experts, indicating insufficient sensitivity to stylistic differences. CLASE-Subj performs comparably to or slightly better than LLM-as-a-judge baselines. CLASE-Mix closely matches human expert dispersion, suggesting appropriate discriminative power for stylistic assessment.

\subsection{Ablation Study}

\begin{figure}
    \centering
    \includegraphics[width=1\linewidth]{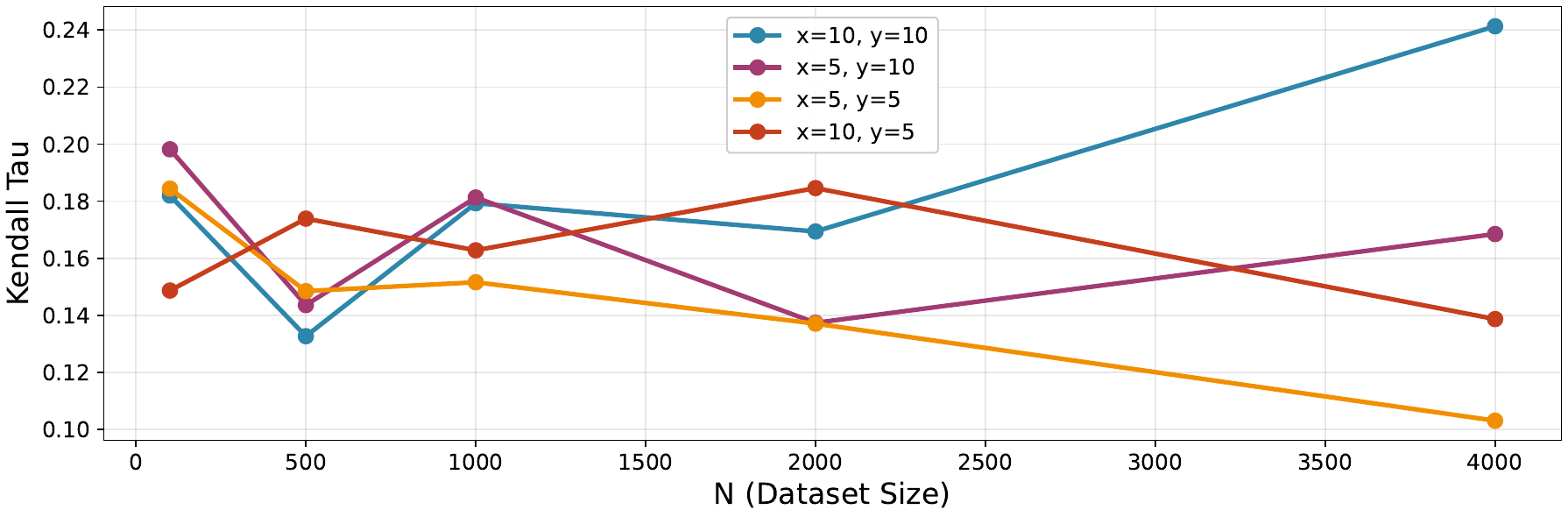}
    \caption{Ablation study on training set size and retrieval parameters for subjective scoring component, measured by Kendall's $\tau$.}
    \label{fig:ablation_subjective}
\end{figure}

\begin{figure}
    \centering
    \includegraphics[width=1\linewidth]{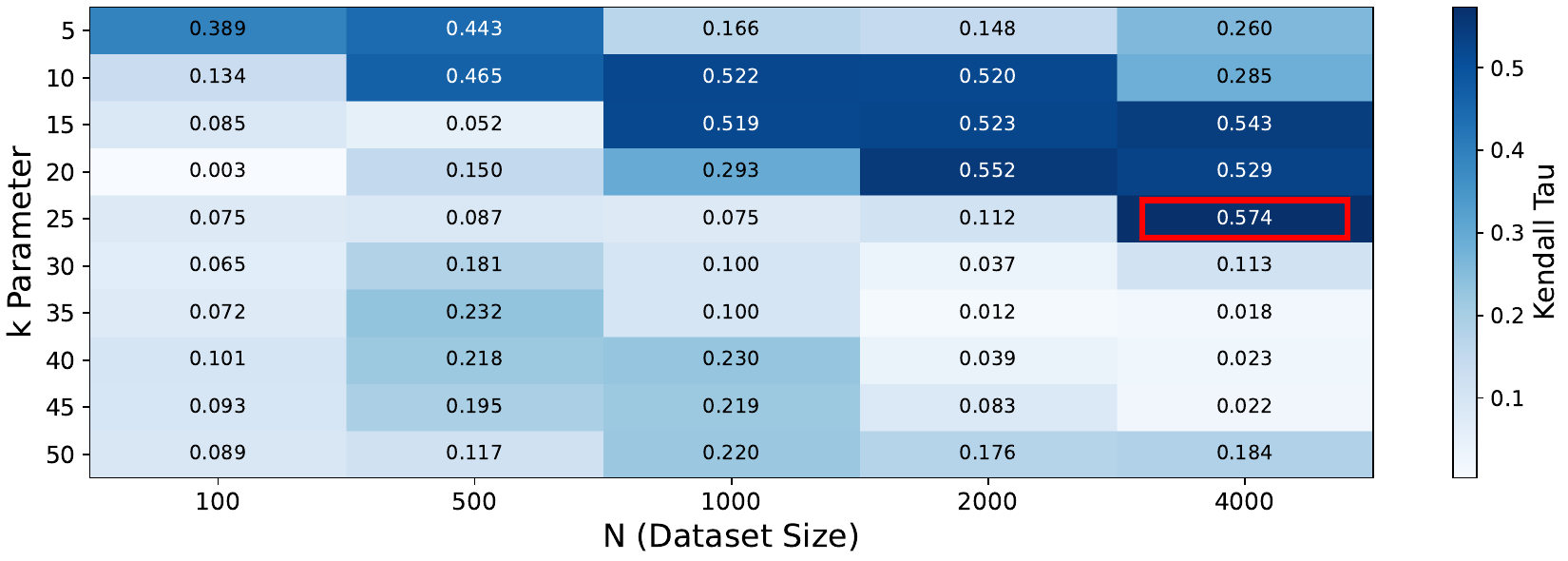}
    \caption{Ablation study on training set size and significant feature count for objective scoring component, measured by Kendall's $\tau$.}
    \label{fig:ablation_objective}
\end{figure}

For the subjective score (Figure \ref{fig:ablation_subjective}), performance generally improves with a larger training set size ($N$), more queries per document ($x$), and more retrieved example pairs per query ($y$). This shows that a richer context of positive and negative examples helps guide the LLM toward more accurate evaluations. For the objective score (Figure \ref{fig:ablation_objective}), we analyzed the impact of training set size ($N$) and the number of significant features ($k$). Correlation peaks at $N=4000, k=25$, indicating that a focused set of salient linguistic markers is optimal.

\subsection{Case Study: Stylistic Feedback}
\label{feedback}

\begin{figure}
    \centering
    \includegraphics[width=1\linewidth]{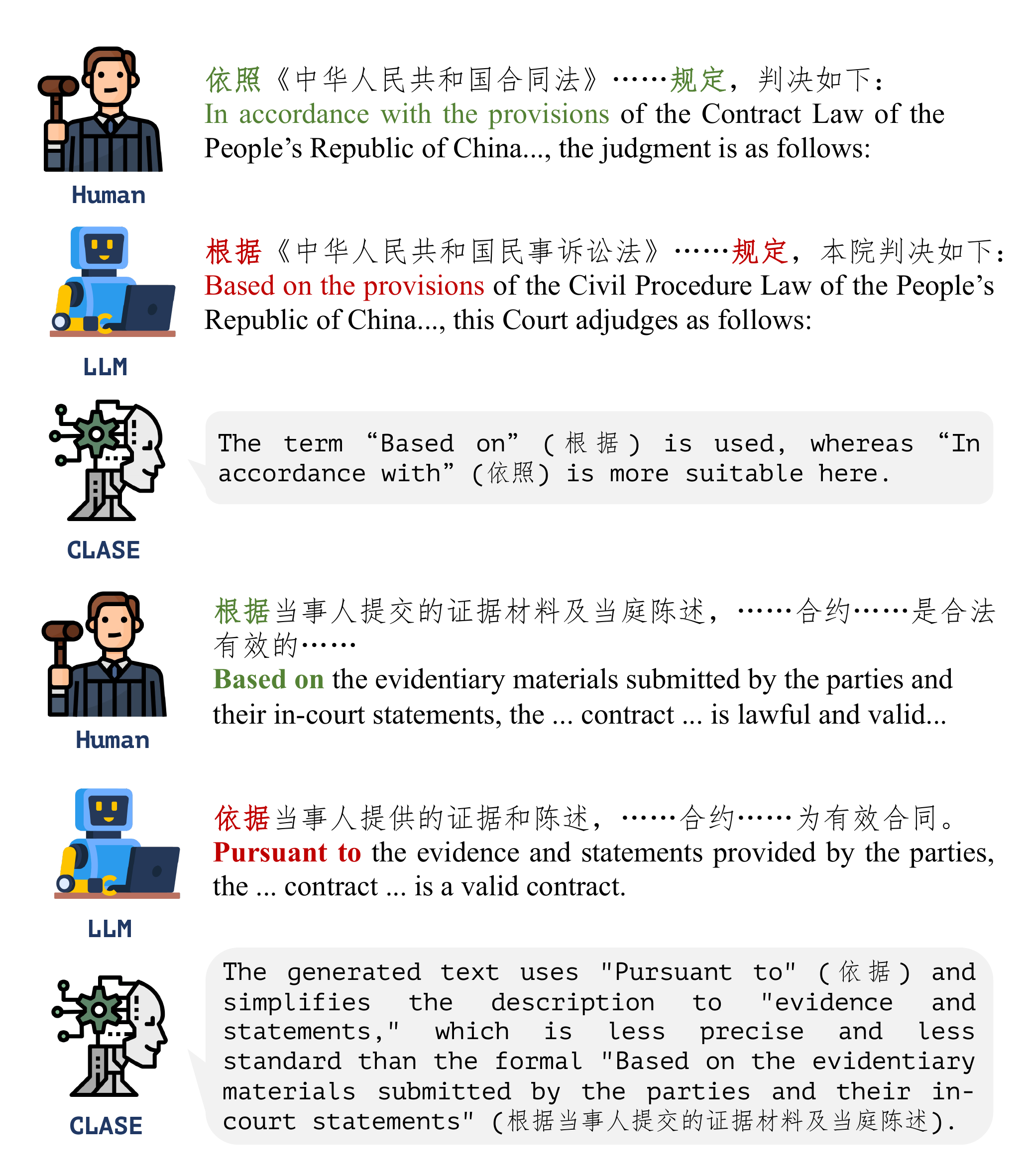}
    \caption{Example of CLASE's natural language feedback identifying specific stylistic issues and improvement suggestions.}
    \label{fig:case}
\end{figure}

Due to preceding CoT reasoning, CLASE-Subj provides interpretable natural language reason/feedback that identifies stylistic deficiencies and improvement directions. Figure \ref{fig:case} demonstrates how CLASE offers actionable comments about legal writing quality. The feedback examples represent issues that practitioners consistently identify when reviewing LLM-generated legal documents in actual judicial settings. Through field observations in Chinese courts, we documented frequent stylistic complaints from legal professionals, including: 

1) inappropriate lexical choices for legal basis versus factual reasoning (different formal terms must be used when citing statutes versus referencing evidentiary materials); 2) omission of conventional emphatic phrases that signal judicial authority and legal compliance (``reasonable and lawful'', ``in accordance with the law''); and 3) subtle distinctions between copular constructions that carry different degrees of formality and certainty in legal contexts.

CLASE detects all these precise issues without any annotation of such domain-specific knowledge in the example in Figure \ref{fig:case}. The system learns to distinguish nuanced professional conventions purely through contrastive analysis, without explicit knowledge injection. It demonstrates that sophisticated domain-specific stylistic expertise, which traditionally requires years of training, could be \textbf{potentially} acquired through automated, contrastive learning, bringing out exciting ``intelligence'' that pure numerical scores cannot offer.

\subsection{Discussion}

\paragraph{Generalizability} Current validation focuses on Chinese civil judgment reasoning sections. However, CLASE's core contrastive learning principles offer promising extensibility. The framework's reliance on natural language knowledge representation suggests potential generalization advantages—stylistic expertise encoded in natural language exhibits transferability compared to numerical features or domain-specific rules. Extension to other legal domains, broader professional writing contexts, or other languages would require domain-specific experience pool construction while preserving the underlying methodology.

\paragraph{Computational considerations} While contrastive learning requires substantial resources for pair generation and experience pool construction, the resulting system operates efficiently in deployment. Once trained, experience pools remain static\footnote{The experience pool can be updated periodically to incorporate new domain knowledge and improve performance over time.} and retrieval can scale well with pre-computed embeddings and indexing. The framework's computational profile is front-loaded during development rather than inference, making it suitable for production environments where training costs amortize over extensive usage.

\paragraph{Scaling characteristics} Our ablation studies reveal complex scaling patterns that challenge simple, ``brute force'' ``scaling law'' assumptions. While performance generally improves with training set size up to $N=4000$, marginal gains diminish beyond certain thresholds. Similarly, optimal feature selection ($k=25$) suggests that excessive linguistic features introduce noise rather than improving discrimination. These findings indicate that effective deployment requires careful hyperparameter optimization rather than naive scaling strategies.

\paragraph{Interpretability and Hybrid Necessity} While CLASE-Obj demonstrates strong performance, reliance on objective features alone is insufficient, especially when we consider evaluation as the reward signal of LLM reinforcement loop. Guiding generation policy with only objective metrics is prone to Goodhart's Law; once specific stylistic markers become explicit optimization targets, models might game the metric without genuine stylistic improvement. Furthermore, numerical feature weights offer limited explanatory power. CLASE-Subj is imperative as it provides: 1) semantic grounding to prevent adversarial overfitting to surface features; and 2) actionable natural language feedback (as shown in Section \ref{feedback}), which is essential for human-in-the-loop workflows. 

\section{Conclusion}

We introduce CLASE, a hybrid evaluation framework that addresses the under-explored challenge of stylistic quality assessment in Chinese legal text generation. By combining objective linguistic feature analysis with experience-guided LLM evaluation, our approach provides a reference-free solution that captures both surface-level patterns and implicit professional conventions in Chinese legal writing.

The key insight underlying CLASE is that professional stylistic expertise can be acquired through contrastive analysis rather than explicit annotation. Our training-free contrastive learning framework successfully extracts meaningful stylistic knowledge from authentic legal documents and their deliberately degraded counterparts, eliminating the need for expensive manual score annotations while maintaining alignment with professional standards.

Experimental results on 200 Chinese legal documents demonstrate that CLASE achieves higher correlation with human expert judgments compared to traditional metrics and pure LLM-based approaches. The framework not only provides quantitative scores but also generates interpretable natural language feedback that identifies specific stylistic deficiencies and improvement directions—capabilities that numerical metrics cannot offer.

The work makes three primary contributions to the intersection of natural language evaluation and legal AI: establishing a hybrid architecture that balances objectivity with nuanced judgment, demonstrating that sophisticated domain-specific conventions can emerge from automated contrastive analysis, and providing a scalable framework that extends beyond Chinese legal text to other domains requiring specialized stylistic conformity.

While current validation focuses on civil judgment documents within a specific model family, the core principles of contrastive stylistic learning offer a general approach to professional text evaluation. Future work should explore cross-domain generalization, computational efficiency optimization, and deeper integration of legal writing principles to enhance both performance and interpretability. As legal AI systems become increasingly sophisticated in generating factually accurate content, frameworks like CLASE become essential for ensuring that automated text generation meets the professional standards expected in legal practice.

\section{Ethical Considerations}

This research primarily involves publicly available legal documents (Chinese civil judgments) and standard large language models. The datasets do not contain private or sensitive individual information that is not already part of the public record. The proposed evaluation framework aims to improve the professional quality of automated legal writing. We anticipate no direct negative societal impacts. However, users should be aware that automated evaluation tools, including CLASE, should serve as assistants to, rather than replacements for, human legal professionals.

\section{Limitations}

We acknowledge several limitations in the current work. First, our experiments are restricted to Chinese civil judgments. While we believe the contrastive learning methodology is transferrable, the specific implementation of the objective component (using Chinese linguistic features) is language-dependent and would require adaptation for other languages. Second, the contrastive pair synthesis relies on the assumption that the degradation model alters style without corrupting semantics. While our strong correlation results suggest the validity of the generated data for training, we did not perform a large-scale manual audit of the synthetic pairs. Third, compared to lightweight metrics like BLEU, CLASE involves a multi-stage pipeline which demands higher computational resources, potentially limiting immediate adoption in low-latency applications.

While CLASE-Obj demonstrates strong empirical performance, the underlying mechanisms by which logistic regression weights capture professional stylistic preferences remain opaque. CLASE-Subj exhibits sophisticated pattern recognition capabilities but lacks deep causal understanding. It identifies specific stylistic deficiencies without comprehending the underlying professional rationales or historical evolution of legal writing conventions. This ``knowing-what but not knowing-why'' limitation restricts the framework's ability to provide educational insights or adapt to evolving professional standards. Future work should explore incorporating explicit legal writing principles to achieve more comprehensive stylistic expertise.

\section{Acknowledgements}
We thank the reviewers for their human touch, dedication, and insightful feedback. This work was supported by the Natural Language Processing Lab at Tsinghua University (TsinghuaNLP).

\bibliographystyle{lrec2026-natbib}
\bibliography{lrec2026-example}

@inproceedings{ma-2025-androids,
    title = "Do Androids Question Electric Sheep? A Multi-Agent Cognitive Simulation of Philosophical Reflection on Hybrid Table Reasoning",
    author = "Ma, Yiran Rex",
    editor = "Zhao, Jin  and
      Wang, Mingyang  and
      Liu, Zhu",
    booktitle = "Proceedings of the 63rd Annual Meeting of the Association for Computational Linguistics (Volume 4: Student Research Workshop)",
    month = jul,
    year = "2025",
    address = "Vienna, Austria",
    publisher = "Association for Computational Linguistics",
    url = "https://aclanthology.org/2025.acl-srw.9/",
    doi = "10.18653/v1/2025.acl-srw.9",
    pages = "143--164",
    ISBN = "979-8-89176-254-1",
}

@article{chan2023chateval,
  title={ChatEval: Towards Better LLM-based Evaluators through Multi-Agent Debate},
  author={Chan, Chi-Min and Chen, Weize and Su, Yujia and Yu, Jiahui and Xue, Wei and Zhang, Shanghang and Fu, Jie and Liu, Zhiyuan},
  journal={arXiv preprint arXiv:2308.07201},
  year={2023}
}

@inproceedings{papineni2002bleu,
  title={Bleu: a method for automatic evaluation of machine translation},
  author={Papineni, Kishore and Roukos, Salim and Ward, Todd and Zhu, Wei-Jing},
  booktitle={Proceedings of the 40th annual meeting of the Association for Computational Linguistics},
  pages={311--318},
  year={2002}
}

@inproceedings{zhang2019bertscore,
title={BERTScore: Evaluating Text Generation with BERT},
author={Tianyi Zhang and Varsha Kishore and Felix Wu and Kilian Q. Weinberger and Yoav Artzi},
booktitle={International Conference on Learning Representations},
year={2020},
url={https://openreview.net/forum?id=SkeHuCVFDr}
}

@inproceedings{lin2004rouge,
  title={Rouge: A package for automatic evaluation of summaries},
  author={Lin, Chin-Yew},
  booktitle={Text summarization branches out},
  pages={74--81},
  year={2004}
}

@article{bommarito2022gpt,
  title={{GPT-4} passes the bar exam},
  author={Katz, Daniel Martin and Bommarito, Michael James and Gao, Shang and Arredondo, Pablo},
  journal={Philosophical Transactions of the Royal Society A},
  volume={382},
  number={2270},
  pages={20230254},
  year={2024},
  publisher={The Royal Society}
}

@book{tiersma1999legal,
  title={Legal Language},
  author={Tiersma, Peter Meijes},
  publisher={University of Chicago Press},
  year={1999}
}

@article{foley2002language,
  title={Language, Law and Legal Writing: An Introduction to Legal Discourse},
  author={Foley, Richard},
  journal={Legal Writing: The Journal of the Legal Writing Institute},
  volume={8},
  pages={1--35},
  year={2002}
}

@inproceedings{chalkidis2022lexglue,
  title={LexGLUE: A benchmark dataset for legal language understanding in English},
  author={Chalkidis, Ilias and Jana, Abhik and Hartung, Dirk and Bommarito, Michael and Androutsopoulos, Ion and Katz, Daniel and Aletras, Nikolaos},
  booktitle={Proceedings of the 60th Annual Meeting of the Association for Computational Linguistics (Volume 1: Long Papers)},
  pages={4310--4330},
  year={2022}
}

@misc{zhong2018cail2018,
      title={CAIL2018: A Large-Scale Legal Dataset for Judgment Prediction}, 
      author={Chaojun Xiao and Haoxi Zhong and Zhipeng Guo and Cunchao Tu and Zhiyuan Liu and Maosong Sun and Yansong Feng and Xianpei Han and Zhen Hu and Heng Wang and Jianfeng Xu},
      year={2018},
      eprint={1807.02478},
      archivePrefix={arXiv},
      primaryClass={cs.CL},
      url={https://arxiv.org/abs/1807.02478}, 
}

@inproceedings{fu2022jec,
  title={JEC-QA: a legal-domain question answering dataset},
  author={Zhong, Haoxi and Xiao, Chaojun and Tu, Cunchao and Zhang, Tianyang and Liu, Zhiyuan and Sun, Maosong},
  booktitle={Proceedings of the AAAI conference on artificial intelligence},
  volume={34},
  number={05},
  pages={9701--9708},
  year={2020}
}

@article{mellish1998evaluation,
  title={Evaluation in the context of natural language generation},
  author={Mellish, Chris and Dale, Robert},
  journal={Computer Speech \& Language},
  volume={12},
  number={4},
  pages={349--373},
  year={1998},
  publisher={Elsevier}
}

@article{wang2023survey,
  title={A Survey on Large Language Model based Autonomous Agents},
  author={Wang, Lei and Ma, Chen and Feng, Xueyang and Zhang, Zeyu and Yang, Hao and Zhang, Jingsen and Chen, Zhiyuan and Tang, Jiakai and Chen, Xu and Lin, Yankai and others},
  journal={arXiv preprint arXiv:2308.11432},
  year={2023}
}

@article{li2020cohesion,
  title={A Study of Cohesion in the Chinese Legal Text: Based on Criminal Procedure Law of the People's Republic of China},
  author={Li, Shifang and Wang, Yifan},
  journal={Theory and Practice in Language Studies},
  volume={11},
  number={12},
  pages={1709--1716},
  year={2021},
  publisher={Academy Publication Co., Ltd.}
}

@article{lu2021legal,
  title={Legal reasoning: A textual perspective on common law judicial opinions and Chinese judgments},
  author={Lu, Nan and Yuan, Chuanyou},
  journal={Text \& Talk},
  volume={41},
  number={1},
  pages={71--93},
  year={2021},
  publisher={De Gruyter}
}

@article{sun2017linguistic,
url = {https://doi.org/10.1515/ijld-2017-0017},
title = {Linguistic variation and legal representation in legislative discourse: A corpus-based multi-dimensional study},
author = {Yuxiu Sun and Le Cheng},
pages = {315--339},
volume = {2},
number = {2},
journal = {International Journal of Legal Discourse},
doi = {doi:10.1515/ijld-2017-0017},
year = {2017},
}

@article{li2022lexical,
  title={Lexical, Syntactic and Textual Features of Shipping Legal Documents in Chinese and English},
  author={Li, Huixian},
  journal={SCIREA Journal of Sociology},
  volume={6},
  number={2},
  pages={83--103},
  year={2022},
  publisher={SCIREA}
}

@book{krippendorff2011computing,
  title={Computing Krippendorff's Alpha Reliability},
  author={Krippendorff, Klaus},
  publisher={University of Pennsylvania},
  year={2011}
}

@misc{qwen25_2024,
      title={Qwen2.5 Technical Report}, 
      author={Qwen and : and An Yang and Baosong Yang and Beichen Zhang and Binyuan Hui and Bo Zheng and Bowen Yu and Chengyuan Li and Dayiheng Liu and Fei Huang and Haoran Wei and Huan Lin and Jian Yang and Jianhong Tu and Jianwei Zhang and Jianxin Yang and Jiaxi Yang and Jingren Zhou and Junyang Lin and Kai Dang and Keming Lu and Keqin Bao and Kexin Yang and Le Yu and Mei Li and Mingfeng Xue and Pei Zhang and Qin Zhu and Rui Men and Runji Lin and Tianhao Li and Tianyi Tang and Tingyu Xia and Xingzhang Ren and Xuancheng Ren and Yang Fan and Yang Su and Yichang Zhang and Yu Wan and Yuqiong Liu and Zeyu Cui and Zhenru Zhang and Zihan Qiu},
      year={2025},
      eprint={2412.15115},
      archivePrefix={arXiv},
      primaryClass={cs.CL},
      url={https://arxiv.org/abs/2412.15115}, 
}

@article{openai2024gpt4o,
  title={GPT-4o System Card},
  author={OpenAI},
  journal={arXiv preprint arXiv:2410.21276},
  year={2024}
}

@inproceedings{kwon2023efficient,
  title={Efficient Memory Management for Large Language Model Serving with PagedAttention},
  author={Kwon, Woosuk and Li, Zhuohan and Zhuang, Siyuan and Sheng, Ying and Zheng, Lianmin and Yu, Cody Hao and Gonzalez, Joseph E. and Zhang, Hao and Stoica, Ion},
  booktitle={Proceedings of the 29th Symposium on Operating Systems Principles},
  year={2023}
}

@article{wei_chain--thought_2023,
  title={Chain-of-Thought Prompting Elicits Reasoning in Large Language Models},
  author={Wei, Jason and Wang, Xuezhi and Schuurmans, Dale and Bosma, Maarten and Ichter, Brian and Xia, Fei and Chi, Ed and Le, Quoc and Zhou, Denny},
  journal={Advances in Neural Information Processing Systems},
  volume={35},
  pages={24824--24837},
  year={2022}
}

@incollection{huang_exploring_2018,
	address = {Cham},
	title = {Exploring the {Impact} of {Linguistic} {Features} for {Chinese} {Readability} {Assessment}},
	volume = {10619},
	isbn = {978-3-319-73617-4 978-3-319-73618-1},
	url = {http://link.springer.com/10.1007/978-3-319-73618-1_67},
	doi = {10.1007/978-3-319-73618-1_67},
	language = {en},
	urldate = {2024-11-29},
	booktitle = {Natural {Language} {Processing} and {Chinese} {Computing}},
	publisher = {Springer International Publishing},
	author = {Qiu, Xinying and Deng, Kebin and Qiu, Likun and Wang, Xin},
	editor = {Huang, Xuanjing and Jiang, Jing and Zhao, Dongyan and Feng, Yansong and Hong, Yu},
	year = {2018},
	pages = {771--783},
}

@article{pearson1895note,
  title={Note on regression and inheritance in the case of two parents},
  author={Pearson, Karl},
  journal={Proceedings of the royal society of London},
  volume={58},
  number={347-352},
  pages={240--242},
  year={1895}
}

@article{spearman1904proof,
  title={The proof and measurement of association between two things},
  author={Spearman, Charles},
  journal={The American journal of psychology},
  volume={15},
  number={1},
  pages={72--101},
  year={1904}
}

@article{kendall1938new,
  title={A new measure of rank correlation},
  author={Kendall, Maurice G},
  journal={Biometrika},
  volume={30},
  number={1/2},
  pages={81--93},
  year={1938}
}

@inproceedings{banerjee2005meteor,
  title={METEOR: An automatic metric for MT evaluation with improved correlation with human judgments},
  author={Banerjee, Satanjeev and Lavie, Alon},
  booktitle={Proceedings of the acl workshop on intrinsic and extrinsic evaluation measures for machine translation and/or summarization},
  pages={65--72},
  year={2005}
}

@inproceedings{zhao2019moverscore,
    title = "{M}over{S}core: Text Generation Evaluating with Contextualized Embeddings and Earth Mover Distance",
    author = "Zhao, Wei  and
      Peyrard, Maxime  and
      Liu, Fei  and
      Gao, Yang  and
      Meyer, Christian M.  and
      Eger, Steffen",
    editor = "Inui, Kentaro  and
      Jiang, Jing  and
      Ng, Vincent  and
      Wan, Xiaojun",
    booktitle = "Proceedings of the 2019 Conference on Empirical Methods in Natural Language Processing and the 9th International Joint Conference on Natural Language Processing (EMNLP-IJCNLP)",
    month = nov,
    year = "2019",
    address = "Hong Kong, China",
    publisher = "Association for Computational Linguistics",
    url = "https://aclanthology.org/D19-1053/",
    doi = "10.18653/v1/D19-1053",
    pages = "563--578",
}

@article{reiter2009validity,
  title={An investigation into the validity of some metrics for automatically evaluating natural language generation systems},
  author={Reiter, Ehud and Belz, Anja},
  journal={Computational Linguistics},
  volume={35},
  number={4},
  pages={529--558},
  year={2009},
  publisher={MIT Press One Rogers Street, Cambridge, MA 02142-1209, USA journals-info~…}
}

@article{novikova2017need,
  title={Why we need new evaluation metrics for NLG},
  author={Novikova, Jekaterina and Du{\v{s}}ek, Ond{\v{r}}ej and Curry, Amanda Cercas and Rieser, Verena},
  journal={arXiv preprint arXiv:1707.06875},
  year={2017}
}

@misc{liu2023geval,
      title={G-Eval: NLG Evaluation using GPT-4 with Better Human Alignment}, 
      author={Yang Liu and Dan Iter and Yichong Xu and Shuohang Wang and Ruochen Xu and Chenguang Zhu},
      year={2023},
      eprint={2303.16634},
      archivePrefix={arXiv},
      primaryClass={cs.CL},
      url={https://arxiv.org/abs/2303.16634}, 
}

@article{schmidt2021large,
  title={Leveraging large language models for nlg evaluation: A survey},
  author={Li, Zhen and Xu, Xiaohan and Shen, Tao and Xu, Can and Gu, Jia-Chen and Tao, Chongyang},
  journal={CoRR},
  year={2024}
}

@inproceedings{rei2021references,
  title={Are references really needed? unbabel-IST 2021 submission for the metrics shared task},
  author={Rei, Ricardo and Farinha, Ana C and Zerva, Chrysoula and Van Stigt, Daan and Stewart, Craig and Ramos, Pedro and Glushkova, Taisiya and Martins, Andr{\'e} FT and Lavie, Alon},
  booktitle={Proceedings of the Sixth Conference on Machine Translation},
  pages={1030--1040},
  year={2021}
}

@article{chalkidis2020legal,
  title={LEGAL-BERT: The muppets straight out of law school},
  author={Chalkidis, Ilias and Fergadiotis, Manos and Malakasiotis, Prodromos and Aletras, Nikolaos and Androutsopoulos, Ion},
  journal={arXiv preprint arXiv:2010.02559},
  year={2020}
}

@misc{zhou2024lawgpt,
      title={LawGPT: A Chinese Legal Knowledge-Enhanced Large Language Model}, 
      author={Zhi Zhou and Jiang-Xin Shi and Peng-Xiao Song and Xiao-Wen Yang and Yi-Xuan Jin and Lan-Zhe Guo and Yu-Feng Li},
      year={2024},
      eprint={2406.04614},
      archivePrefix={arXiv},
      primaryClass={cs.CL},
      url={https://arxiv.org/abs/2406.04614}, 
}

@article{cui2023chatlaw,
  title={Chatlaw: Open-source legal large language model with integrated external knowledge bases},
  author={Cui, Jiaxi and Li, Zongjian and Yan, Yang and Chen, Bohua and Yuan, Li},
  journal={CoRR},
  year={2023}
}

@misc{huang2023lawyer,
      title={Lawyer LLaMA Technical Report}, 
      author={Quzhe Huang and Mingxu Tao and Chen Zhang and Zhenwei An and Cong Jiang and Zhibin Chen and Zirui Wu and Yansong Feng},
      year={2023},
      eprint={2305.15062},
      archivePrefix={arXiv},
      primaryClass={cs.CL},
      url={https://arxiv.org/abs/2305.15062}, 
}

@misc{li2025legaleval,
      title={{LegalEval-Q}: A New Benchmark for The Quality Evaluation of LLM-Generated Legal Text}, 
      author={Li yunhan and Wu gengshen},
      year={2025},
      eprint={2505.24826},
      archivePrefix={arXiv},
      primaryClass={cs.CL},
      url={https://arxiv.org/abs/2505.24826}, 
}

@inproceedings{dai2023laiw,
  title={{LAiW}: A {Chinese} legal large language models benchmark},
  author={Dai, Yongfu and Feng, Duanyu and Huang, Jimin and Jia, Haochen and Xie, Qianqian and Zhang, Yifang and Han, Weiguang and Tian, Wei and Wang, Hao},
  booktitle={Proceedings of the 31st International conference on computational linguistics},
  pages={10738--10766},
  year={2025}
}

@misc{hou2024gaps,
      title={Gaps or Hallucinations? Gazing into Machine-Generated Legal Analysis for Fine-grained Text Evaluations}, 
      author={Abe Bohan Hou and William Jurayj and Nils Holzenberger and Andrew Blair-Stanek and Benjamin Van Durme},
      year={2024},
      eprint={2409.09947},
      archivePrefix={arXiv},
      primaryClass={cs.CL},
      url={https://arxiv.org/abs/2409.09947}, 
}

@article{deroy2023ready,
  title={How ready are pre-trained abstractive models and LLMs for legal case judgement summarization?},
  author={Deroy, Aniket and Ghosh, Kripabandhu and Ghosh, Saptarshi},
  journal={arXiv preprint arXiv:2306.01248},
  year={2023}
}

@misc{savelka2023explaining,
      title={Explaining Legal Concepts with Augmented Large Language Models (GPT-4)}, 
      author={Jaromir Savelka and Kevin D. Ashley and Morgan A. Gray and Hannes Westermann and Huihui Xu},
      year={2023},
      eprint={2306.09525},
      archivePrefix={arXiv},
      primaryClass={cs.CL},
      url={https://arxiv.org/abs/2306.09525}, 
}

@article{guha2023legalbench,
  title={Legalbench: A collaboratively built benchmark for measuring legal reasoning in large language models},
  author={Guha, Neel and Nyarko, Julian and Ho, Daniel and R{\'e}, Christopher and Chilton, Adam and Chohlas-Wood, Alex and Peters, Austin and Waldon, Brandon and Rockmore, Daniel and Zambrano, Diego and others},
  journal={Advances in neural information processing systems},
  volume={36},
  pages={44123--44279},
  year={2023}
}

@misc{shao2025legal,
      title={When Large Language Models Meet Law: Dual-Lens Taxonomy, Technical Advances, and Ethical Governance}, 
      author={Peizhang Shao and Linrui Xu and Jinxi Wang and Wei Zhou and Xingyu Wu},
      year={2025},
      eprint={2507.07748},
      archivePrefix={arXiv},
      primaryClass={cs.CL},
      url={https://arxiv.org/abs/2507.07748}, 
}

@article{juola2006authorship,
  title={Authorship attribution},
  author={Juola, Patrick and others},
  journal={Foundations and Trends{\textregistered} in Information Retrieval},
  volume={1},
  number={3},
  pages={233--334},
  year={2008},
  publisher={Now Publishers, Inc.}
}

@article{argamon2003stylistic,
  title={Stylistic text classification using functional lexical features},
  author={Argamon, Shlomo and Whitelaw, Casey and Chase, Paul and Hota, Sobhan Raj and Garg, Navendu and Levitan, Shlomo},
  journal={Journal of the American Society for Information Science and Technology},
  volume={58},
  number={6},
  pages={802--822},
  year={2007},
  publisher={Wiley Online Library}
}

@inproceedings{styledistance2024,
    title = "{S}tyle{D}istance: Stronger Content-Independent Style Embeddings with Synthetic Parallel Examples",
    author = "Patel, Ajay  and
      Zhu, Jiacheng  and
      Qiu, Justin  and
      Horvitz, Zachary  and
      Apidianaki, Marianna  and
      McKeown, Kathleen  and
      Callison-Burch, Chris",
    editor = "Chiruzzo, Luis  and
      Ritter, Alan  and
      Wang, Lu",
    booktitle = "Proceedings of the 2025 Conference of the Nations of the Americas Chapter of the Association for Computational Linguistics: Human Language Technologies (Volume 1: Long Papers)",
    month = apr,
    year = "2025",
    address = "Albuquerque, New Mexico",
    publisher = "Association for Computational Linguistics",
    url = "https://aclanthology.org/2025.naacl-long.436/",
    doi = "10.18653/v1/2025.naacl-long.436",
    pages = "8662--8685",
    ISBN = "979-8-89176-189-6",
}

@inproceedings{fei2023lawbench,
  title={Lawbench: Benchmarking legal knowledge of large language models},
  author={Fei, Zhiwei and Shen, Xiaoyu and Zhu, Dawei and Zhou, Fengzhe and Han, Zhuo and Huang, Alan and Zhang, Songyang and Chen, Kai and Yin, Zhixin and Shen, Zongwen and others},
  booktitle={Proceedings of the 2024 conference on empirical methods in natural language processing},
  pages={7933--7962},
  year={2024}
}

@inproceedings{gan2025ucl,
  title={UCL-Bench: A Chinese User-Centric Legal Benchmark for Large Language Models},
  author={Gan, Ruoli and Feng, Duanyu and Zhang, Chen and Lin, Zhihang and Jia, Haochen and Wang, Hao and Cai, Zhenyang and Cui, Lei and Xie, Qianqian and Huang, Jimin and others},
  booktitle={Findings of the Association for Computational Linguistics: NAACL 2025},
  pages={7945--7988},
  year={2025}
}

@misc{su2025judge,
      title={{JuDGE}: Benchmarking Judgment Document Generation for Chinese Legal System}, 
      author={Weihang Su and Baoqing Yue and Qingyao Ai and Yiran Hu and Jiaqi Li and Changyue Wang and Kaiyuan Zhang and Yueyue Wu and Yiqun Liu},
      year={2025},
      eprint={2503.14258},
      archivePrefix={arXiv},
      primaryClass={cs.CL},
      url={https://arxiv.org/abs/2503.14258}, 
}

@article{li2025casegen,
  title={CaseGen: A Benchmark for Multi-Stage Legal Case Documents Generation},
  author={Li, Haitao and Ye, Jiaying and Hu, Yiran and Chen, Jia and Ai, Qingyao and Wu, Yueyue and Chen, Junjie and Chen, Yifan and Luo, Cheng and Zhou, Quan and Liu, Yiqun},
  journal={arXiv preprint arXiv:2502.17943},
  year={2025}
}

@misc{courtWritingGuide2010,
  title={Court Writing Guide},
  author={{Court Writing Committee}},
  year={2010},
  note={Legal writing guidelines for professional practice}
}

@inproceedings{zheng_judging_2023,
title={Judging {LLM}-as-a-Judge with {MT}-Bench and Chatbot Arena},
author={Lianmin Zheng and Wei-Lin Chiang and Ying Sheng and Siyuan Zhuang and Zhanghao Wu and Yonghao Zhuang and Zi Lin and Zhuohan Li and Dacheng Li and Eric Xing and Hao Zhang and Joseph E. Gonzalez and Ion Stoica},
booktitle={Thirty-seventh Conference on Neural Information Processing Systems Datasets and Benchmarks Track},
year={2023},
url={https://openreview.net/forum?id=uccHPGDlao}
}

@inproceedings{li_leveraging_2024,
  title={Leveraging large language models for learning complex legal concepts through storytelling},
  author={Jiang, Hang and Zhang, Xiajie and Mahari, Robert and Kessler, Daniel and Ma, Eric and August, Tal and Li, Irene and Pentland, Alex and Kim, Yoon and Roy, Deb and others},
  booktitle={Proceedings of the 62nd Annual Meeting of the Association for Computational Linguistics (Volume 1: Long Papers)},
  pages={7194--7219},
  year={2024}
}

@misc{ito_reference-free_2025,
      title={Reference-free Evaluation Metrics for Text Generation: A Survey}, 
      author={Takumi Ito and Kees van Deemter and Jun Suzuki},
      year={2025},
      eprint={2501.12011},
      archivePrefix={arXiv},
      primaryClass={cs.CL},
      url={https://arxiv.org/abs/2501.12011}, 
}

@article{jin_deep_2022,
  title={Deep learning for text style transfer: A survey},
  author={Jin, Di and Jin, Zhijing and Hu, Zhiting and Vechtomova, Olga and Mihalcea, Rada},
  journal={Computational Linguistics},
  volume={48},
  number={1},
  pages={155--205},
  year={2022},
  publisher={MIT Press One Broadway, 12th Floor, Cambridge, Massachusetts 02142, USA~…}
}

@article{bitton2025detectingstylisticfingerprintslarge,
  title={Detecting Stylistic Fingerprints of Large Language Models},
  author={Bitton, Yehonatan and Bitton, Elad and Nisan, Shai},
  journal={arXiv preprint arXiv:2503.01659},
  year={2025}
}

\end{document}